\documentclass[11pt,a4paper]{article}
\usepackage[hyperref]{eacl2021}
\usepackage{times}
\usepackage{latexsym}
\usepackage{graphicx}
\usepackage{multirow}

\newcommand{\red}[1]{\textcolor{red}{#1}}
\newcommand{\hide}[1]{}

\usepackage{microtype}

\aclfinalcopy 

\title{Congolese Swahili Machine Translation for Humanitarian Response}

\author{Alp \"Oktem, Eric DeLuca, Rodrigue Bashizi, Eric Paquin, Grace Tang \\
Translators without Borders\\
\texttt{\{alp,eric,rodrigue,ericpaquin,grace\}@translatorswb.org} \\
}

\begin{document}
\maketitle
\begin{abstract}
In this paper we describe our efforts to make a bidirectional Congolese Swahili (SWC) to French (FRA) neural machine translation system with the motivation of improving humanitarian translation workflows. For training, we created a 25,302-sentence general domain parallel corpus and combined it with publicly available data. Experimenting with low-resource methodologies like cross-dialect transfer and semi-supervised learning, we recorded improvements of up to 2.4 and 3.5 BLEU points in the SWC–FRA and FRA–SWC directions, respectively. We performed human evaluations to assess the usability of our models in a COVID-domain chatbot that operates in the Democratic Republic of Congo (DRC). Direct assessment in the SWC–FRA direction demonstrated an average quality ranking of 6.3 out of 10 with 75\% of the target strings conveying the main message of the source text. For the FRA–SWC direction, our preliminary tests on post-editing assessment showed its potential usefulness for machine-assisted translation. We make our models, datasets containing up to 1 million sentences, our development pipeline, and a translator web-app available for public use. 

\end{abstract}

\section{Introduction}

Swahili (Kiswahili among its speakers) is a macrolanguage spoken widely in east Africa with an estimated  100 to 150 million speakers. It is the official language in Tanzania and Kenya, where it is referred to as coastal Swahili, and one of four national languages in the Democratic Republic of Congo (DRC), where it is referred to as Congolese Swahili. Coastal and Congolese Swahili differ substantially in terms of vocabulary, grammar, and structure. This is largely due to educational politics and differences in the colonial languages of the countries they are spoken in. For example, compared to coastal Swahili, Congolese Swahili has a lot of French and Lingala influence. To illustrate, take the translations of ``Once the test is negative, the family can take care of the funeral themselves'':

\begin{description}
\item [Coastal Swahili] {\it Mara tu upimaji ukiwa hasi, familia inaweza kushughulikia mazishi wao wenyewe.} 
\item [Congolese Swahili] {\it Ikiwa tu vipimovinaonesha kama ni mtu ambaye hakuhakikishwa ku kuwa na ugonjwa, familia inaweza kufanya mazishi yenyewe.}
\end{description}

The translations differ in vocabulary and structure. In the Congolese dialect, the first clause needs to be expressed as “If the test shows that someone is not sick,” since a commonly used word for “negative” (\textit{Hasi} in the coastal dialect) does not exist. The conditionality is expressed with \textit{Ikiwa} (“if”) instead of \textit{Mara} (“once”), as the latter is very rarely used in DRC. And finally, the Congolese Swahili translation uses the colloquial word \textit{kufanya} for “dealing with a funeral” instead of \textit{kushughulikia} (“care”), since "taking care of a funeral" does not mean anything in common speech. 

This example hints that machine translation (MT) engines that are solely based on the coastal dialect could be ineffective in delivering or receiving sensitive information. That is why there is a need for dialect-specific MT development to aid humanitarian relief efforts in DRC.

Translators without Borders (TWB) specializes in helping people affected by crisis to get information in their language in a format they understand. Communicating effectively in the languages and formats people understand is central to ensuring that people understand health risks and know how to keep themselves and their families safe. TWB translates vital messages into local languages and works with responders to develop tools and language capacity to provide communities with better access to information and services that meet their needs. TWB has been active in eastern DRC since early 2019 with the Ebola response. In October 2020, TWB launched a multilingual COVID-19 chatbot that leverages natural language understanding to allow users to ask questions in their own words and receive relevant answers in the same language.

Language technology, such as machine translation, plays an important role in crisis response, increasing the capacity to communicate critical information and key messages in the languages people understand at speed and at scale~\citep{DBLP:conf/wmt/LewisMV11}. Crisis-affected people can access content in local languages firsthand through various channels such as websites, news sources, and social media. In the reverse direction, their questions and feedback can be used to inform aid programs to analyze trends and better meet affected people's needs~\citep{GHTC}. For marginalized languages where there are limited translation resources, MT can also help standardize terms and improve translation quality.

The task of creating MT specialized in the Congolese dialect is especially challenging due to the low-resource status of the language. In~\citet{Joshi}’s index that ranks languages in terms of resources, it is listed as “\textit{Left Behind}”. According to their definition, it is substantially difficult to establish a digital foundation for the language with the existing data resources. If the bottleneck of data scarcity is not addressed, it is impossible to develop useful tools such as MT that can greatly serve the translators of a language. In fact, this does not only put translators of this language at a great disadvantage, but it could also contribute to the deprioritization of the language by its speakers both digitally and socially. 

The objectives of this work are to: 

\begin{itemize}
	\item Curate research and public data sources for use in MT development for Congolese Swahili (Section~\ref{sec:related_work}).
	\item Describe the datasets that we created and compiled for this work (Section~\ref{sec:data}).
	\item Investigate the use of low-resource methodologies to deliver optimal bidirectional MT models (Section~\ref{sec:model}).
	\item Assess the usability of the systems for humanitarian response (Section~\ref{sec:results}). 

\end{itemize}

\section{MT and parallel data efforts in Swahili}
\label{sec:related_work}

The only MT-related work that specifically includes the Congolese Swahili dialect is by the grassroots research initiative \textit{Masakhane}~\citep{nekoto2020participatory}. Their paper brings to light how much and why African languages are left behind in terms of natural language processing (NLP) research. Their work addresses this by publicly releasing models and benchmarks for many African languages, including Congolese Swahili. The main data resource used in this work that makes the breadth of benchmarks possible is the JW300 collection~\citep{agic-vulic-2019-jw300}, a parallel corpus of more than 300 languages, including many marginalized ones.

According to Ethnologue\footnote{\url{https://www.ethnologue.com/language/swh} (Subscription required)}, Swahili is a Bantu language spoken by up to 150 million people throughout eastern Africa including in Tanzania, Kenya, Uganda, Rwanda, Burundi, Mozambique, Somali, and DRC. It is comparatively well represented by commercial MT service providers such as Google Translate and Microsoft Translator.

Research on machine translation of Swahili has been taking place for over 50 years. \citet{DBLP:journals/mtcl/Woodhouse68} analysed morphological structure of the language with the aim of building mechanical translation. Mechanical translation logic consists of parsing the source text and mapping each morphological unit to its translation using dictionaries. Even though the mechanical process is outdated in the era of neural machine translation, it still sheds light on the difficulties that any computational translation setup might face. The paper points out the extensive use of prefixes and suffixes in Swahili: subject and object, tense, and negation are all embedded in one word as prefixes, whereas passive, causative, prepositional, reciprocal, subjunctive, plural imperative, and some singular imperative forms are all formed using suffixes. 

\citet{DBLP:journals/lre/PauwWS11} pioneered statistical machine translation in Swahili. Their work presents a 2-million-word parallel corpus paired with English and also part-of-speech tagging annotation from English into Swahili. Their SAWA corpus is not openly distributed, however it is available on demand~\citep{DBLP:conf/eamt/Sanchez-Martinez20}. 

For its large web presence and status as a lingua franca in many countries, Swahili is listed as a “\textit{Rising Star}” in \citet{Joshi}’s index. Recent years have seen an expansion in language coverage for neural machine translation in many low-resource languages. \citet{DBLP:conf/eamt/Sanchez-Martinez20} presents bidirectional English-Swahili MT systems with the motivation of supporting international media outlets that publish in the language. They also address the data scarcity problem by openly publishing crawled monolingual and parallel data. Another recent work by \citet{DBLP:journals/corr/abs-2003-14402} investigates the use of various methodologies like semi-supervised, transfer-learning, and multilingual modeling for building NMT benchmarks for five east African languages: Swahili, Amharic, Tigrigna, Oromo, and Somali, all paired with English. 

\begin{table*}
\centering
\begin{tabular}{llllll}
\hline
\textbf{Dataset id.} & \textbf{Language} & \textbf{Pair} & \textbf{
\#Sentences} & \textbf{\#Tokens} & \textbf{Domain}\\
\hline
TWBkits.swc & SWC & FR & 25,302 & 486,278 & General \\
TICO19.swc & SWC & FR & 3,048 & 159,315 & Medical \\
TWBinTM.swc & SWC & FR & 4,295 & 135,595 & Humanitarian\\
JW300.swc & SWC & FR & 576,110 & 19,724,83 & Bible/General\\
TWBinTM.sw & SW & FR & 462 & 18,960 & Humanitarian\\
GlobalVoices & SW & FR & 17,231 & 672,643 & Bible\\
JW300.sw & SW & FR & 955,776 & 34,217,842 & Bible/General\\
Tanzil & SW & FR & 10,258 & 269,945 & Quran\\
TED2020 & SW & FR & 9,606 & 302,339 & General\\
Wikimatrix & SW & FR & 19,909 & 548,974 & General\\
Multiparacrawl & SW & FR & 50,954 & 2,177,957 & General \\
TICO19.sw & SW & EN & 3,048 & 158,922 & Medical \\
TWBkits.sw & SW & EN & 5,000 & 89,137 & General \\
ELRC & SW & EN & 607 & 24,935 & General \\
GoURMET & SW & EN & 156,061 & 6,316,585 & General \\
\hline
\textbf{Parallel total} &  &  & 1,834,619 & 65,722,290 & \\
\hline
Wikipedia.sw & SW & & 151,756 & 2,891,920 & General\\
WMT-News & SW & & 614,642 & 12,497,567 & News\\
\hline
\textbf{Monolingual total} & & & 766,398 & 15,389,487\\
\hline
\end{tabular}
\caption{\label{table:source_data}
Parallel and monolingual corpora statistics for Congolese and coastal Swahili. All externally sourced parallel datasets are available through OPUS repository~\citep{TIEDEMANN12.463, Christodouloupoulos2015, agic-vulic-2019-jw300, DBLP:conf/eamt/Sanchez-Martinez20, DBLP:journals/corr/abs-1907-05791}.}
\end{table*}

\section{Data description}
\label{sec:data}

Parallel data is the main ingredient to train MT systems. Also referred to as bitext, this is essentially a set of sentences with their translations. To build a bidirectional Congolese Swahili and French model, we needed access to translated data in these languages. We sourced hand-crafted parallel data in this language pair from three sources.

\begin{description}
	\item [Gamayun kits] are a starting point for developing audio and text corpora for languages with few or no pre-existing language data resources. Source sentences for this corpora are selected from Tatoeba corpus\footnote{\url{https://tatoeba.org/}} ensuring they represent everyday language without any domain specificity \citep{GHTC}. The portion published with this work is a set of 25,302 French sentences translated into Congolese Swahili.
	\item [TICO-19 translation memories] are collected to assist translators and build MT benchmarks in 36 languages. Each set consists of 3,071 sentences in COVID-19 domain with quality-checked translations~\citep{DBLP:conf/emnlp/AnastasopoulosC20}.
	\item [TWB in-house translation memories] are collected from various translations that were made within TWB. Most of the source content is from translation requests from NGOs that assist in humanitarian relief in DRC.
\end{description}

The only large external open resource we have found specifically for Congolese Swahili is the JW300 corpus~\citep{agic-vulic-2019-jw300}. Additionally, we decided to collect data in non-dialect Swahili since it would enable us to do transfer learning~\citep{zoph-etal-2016-transfer}. We sourced sentences paired with French and English. All internally sourced and publicly available parallel data sources are listed in Table~\ref{table:source_data}. 

\subsection{Synthetic data creation}

The \textit{TWBkits.sw}, \textit{ELRC} and \textit{GoURMET} datasets are in coastal Swahili paired with English. With a significant total of 161,668 sentences, we decided to include them in our experiments. To use them to train our models paired with French, we used an off-the-shelf machine translation model to translate the English sentences to French. For this, we used an open source model by Helsinki-NLP research group provided through the \textit{huggingface} library\footnote{\url{https://huggingface.co/Helsinki-NLP}}~\citep{DBLP:conf/eamt/TiedemannT20}.

One other technique to improve MT is semi-supervised training~\citep{sennrich-etal-2016-neural}. This process involves expanding the training data with monolingual data paired with their back-translation. For our experiments, we automatically translated 766,398 monolingual Swahili sentences sourced from Wikipedia\footnote{\url{http://kevindonnelly.org.uk}} and News sites~\citep{barrault-etal-2019-findings} to French using a model that was trained during this work using the rest of the available data. All monolingual data sources are also listed in Table~\ref{table:source_data}.

\subsection{Test Data}

We used five test sets to perform automatic and human evaluations of our models: 1000 sentences from Gamayun kits, 500 sentences from the TICO-19 set, 2,478 sentences from the JW300 test set used in \citet{nekoto2020participatory}, 100 user-submitted messages from chatbot conversations and 10 chatbot response strings. 
The first two sets of sentences were randomly sampled from the datasets they belong to. Details on testing datasets are listed in Table~\ref{table:test_data}. 

\begin{table}
\centering
\begin{tabular}{lllll}
\hline
\textbf{Testset id.} & \textbf{Lang.} & \textbf{
\#Seg.} & \textbf{Domain}\\
\hline
TWBkits & SWC  & 1,000 & General \\
TICO19 & SWC  & 500 & Medical \\
JW300 & SWC  & 2,478 & General/Bible \\
Uji-user & SWC  & 100 & COVID-19 \\
Uji-content & FRA  & 10  & COVID-19 \\
\hline
\end{tabular}
\caption{\label{table:test_data}
Test data used for automatic and human evaluations. 
}
\end{table}

\subsection{Reproducibility}

We are publishing both the hand-crafted and synthetically generated parallel data used in this work together with the model weights created from them. They are accessible through our project portal: \url{https://gamayun.translatorswb.org/data/}. 

Researchers who would like to reproduce our work can also find our development pipeline and test sets under the project’s github repository: \url{ https://github.com/translatorswb/TWB-MT/tree/swc-fra-bidirectional}.

\section{Model development}
\label{sec:model}

We followed a three-stage approach for training our models. Each stage is associated with a different dataset mixture (listed in Table~\ref{table:data_mixes}). In the first stage, we obtain a base model by pre-training on all available data: authentic and synthetic, in-domain and out-of-domain, and in-dialect and non-dialect. To assess the effect of adding different datasets, we prepared three intermediate data mixtures: \textit{mix.sw} containing non-dialect parallel data, \textit{mix.mted} containing data that was pair-converted from SW–EN datasets and finally \textit{mix.mono} containing back-translated data generated from monolingual Swahili corpora. In the second stage, we fine-tune the model into the Congolese dialect with both hand-crafted and crawled authentic data (\textit{mix.swc}). In the third stage, we fine-tune the model using our datasets consisting only of hand-crafted translations (\textit{mix.inTWB}). 

\begin{table}
\centering
\begin{tabular}{p{1.2cm}p{3.8cm}p{1.3cm}}
\hline
\textbf{Corpus id.} & \textbf{Datasets in corpus} & \textbf{\#Sent. (clean)} \\
\hline
mix.in & TWBkits.swc*, TWBinTM.swc, TICO19.swc & 28,960  \\
mix.swc & JW300.swc, \textit{mix.in} & 547,796  \\
mix.sw & GlobalVoices, Multiparacrawl, JW300.sw, Tanzil, TWBinTM.sw,  TED, Wikimatrix, TICO19.sw, \textit{mix.swc} & 1,124,440   \\
mix.mted  & ELRC, TWBkits.sw, GoURMET, \textit{mix.sw} & 1,234,069  \\
mix.mono  & WMT-News, Wiki.sw, \textit{mix.mted} & 1,837,632 \\
\hline
\end{tabular}
\caption{\label{table:data_mixes}
Training corpora mixes involved at different training stages. Every dataset builds on the one in the previous row. Cleaning process excludes duplicate, empty, test, and development sentences from the mixes. 
}
\end{table}

\begin{figure}
\centering
\includegraphics[width=0.8\linewidth]{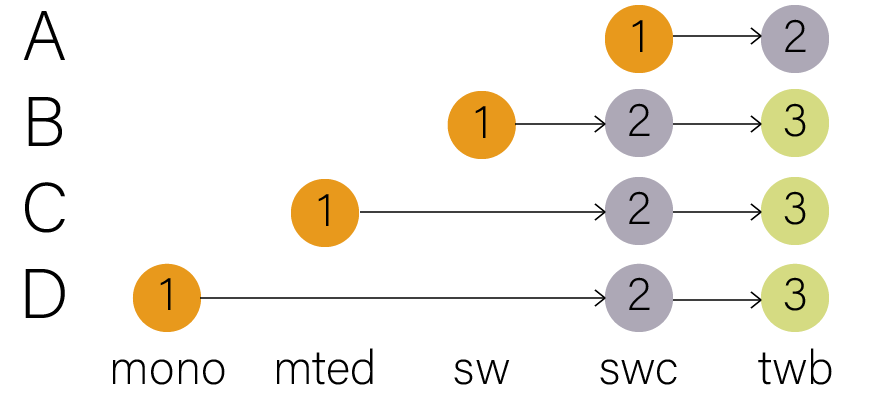}
\caption{Training procedures (A-D) and the corpus mixtures used at each stage (1-3).}
\label{fig:procedures}
\end{figure}

Figure~\ref{fig:procedures} illustrates our training procedures. In procedure A (baseline) we train only on the Congolese Swahili data (mix.swc) and then fine-tune on mix.in. Procedure B performs cross-dialect transfer by pre-training on non-dialect swahili data (mix.sw). Procedure C is similar to Procedure B with addition of pair-converted synthetic data (mix.mted). Finally, Procedure D utilizes synthetic data generated from back-translated monolingual data (mix.mono). In all procedures, we used a validation set of 1000 sentences allocated from mix.in. 

We used the OpenNMT-py toolkit~\citep{klein-etal-2018-opennmt} to train the models. The model consists of an eight-head Transformer~\citep{DBLP:conf/nips/VaswaniSPUJGKP17} with six-layer hidden units of 512 unit size. A token-batch size of 2,048 was selected for multi-dialect and in-dialect training and 512 for final fine-tuning stage. Adam optimizer~\citep{DBLP:journals/corr/KingmaB14} was selected with 4,000 warm-up steps. Trainings were performed until no further improvement was recorded in development set perplexity in the last five validations.

\section{Results}
\label{sec:results}

\subsection{Automatic evaluation}

We report our test set BLEU-scores~\citep{papineni-etal-2002-bleu} for all training procedures and for each stage in Table~\ref{table:bleu}. The results show that cross-dialect transfer learning improves the quality of the models in both directions and in all domains. A pre-training step with non-dialect Swahili data resulted in an increase of 1.5/1.8 BLEU points in TWBkits and 1.7/2 BLEU points in TICO-19 test sets in the SWC–FRA/FRA–SWC directions respectively. In the same test sets, augmenting the pre-training data with synthetically created training data also improved in general. We recorded a 0.4 point increase in TWBkits set in the FRA–SWC direction. In TICO-19 test set, we initially saw a 0.4 BLEU points increase in the SWC–FRA and a 0.7 BLEU points increase in the FRA–SWC directions with the addition of pair-converted corpus (mix.mted). These improvements further increased to 0.7 and 1.5 when we used the back-translated monolingual set (mix.mono). Adding that to the improvement recorded from cross-dialect transfer, TICO-19 set shows greatest performance boost with 2.4 and 3.5 BLEU point increases in the SWC–FRA and FRA–SWC directions. 

For the JW300 test set, we consistently saw a quality decrease in the third training stage. This is because the final fine-tuning stage used data from other sets and domains. Therefore, we evaluate improvements between procedures using the penultimate stage results. Using cross-dialect transfer, we saw 1.8 BLEU points increase in the SWC–FRA direction and 1.4 in the FRA–SWC direction. We can also notice that use of synthetic data only contributed in the SWC–FRA direction with a 1 BLEU point increase between best performing models of procedures B and C. 

Best scoring models for the three sets given in the order of SWC–FRA and FRA–SWC are as follows: 29.2 and 16.6 BLEU points for TWBkits set, 20.1 and 16.5 BLEU points for TICO-19 set, 30.7 and 23.5 BLEU points for JW300 set. The only baseline comparable to previous literature among these is the FRA–SWC direction in the JW300 set. \citet{nekoto2020participatory} reported a BLEU point benchmark of 33.7 using JW300 dataset both for training and testing. The difference in results shows how much multidomain pre-training affects model performance for domain-specific test sets. 

\begin{table*}
\centering
\begin{tabular}{rrr|cc|cc|cc}
 & Dataset  & \multicolumn{1}{l}{} & \multicolumn{2}{c|}{\textbf{TWBkits}}   & \multicolumn{2}{c|}{\textbf{TICO-19}}  & \multicolumn{2}{c}{\textbf{JW300}}     \\
 & \multicolumn{1}{l}{Target} & \multicolumn{1}{l}{} & FRA  & SWC  & FRA  & SWC  & FRA  & SWC  \\ 
\cline{2-9}
\parbox[t]{2mm}{\multirow{11}{*}{\rotatebox[origin=c]{90}{Procedure / Stages}}} & \multirow{2}{*}{A} & 1  & 24.3 & 12.6 & 15.4 & 10.6 & 27.9 & 22.1 \\
 & & 2  & 27.7 & 14.4 & 17.7 & 13   & 26.7 & 19.4  \\ 
\cline{4-9}
 & \multirow{3}{*}{B} & 1  & 24.7 & 15.1 & 17.5 & 12.3 & 29.6 & 20.2 \\
 & & 2  & 24.9 & 15.3 & 17.6 & 13.3 & 29.7 & \textbf{23.5} \\
 & & 3  & \textbf{29.2} & 16.2 & 19.4 & 15   & 28.2 & 20.6\\ 
\cline{4-9}
 & \multirow{3}{*}{C} & 1  & 26   & 14.5 & 17.9 & 12.1 & \textbf{30.7} & 19.9  \\
 & & 2  & 26.2 & 15.5 & 18.1 & 13.3 & 29.8 & 23.2  \\
 & & 3  & \textbf{29.2} & \textbf{16.6} & 19.8 & 15.7 & 29.8 & 21.7 \\ 
\cline{4-9}
 & \multirow{3}{*}{D} & 1  & 24.9 & 15   & 16.8 & 13.1 & 28.7 & 18.8\\
 & & 2  & 26.2 & 15.7 & 17.9 & 14.2 & 29.3 & 22.3   \\
 & & 3  & 28.9 & \textbf{16.6} & \textbf{20.1} & \textbf{16.5} & 28.7 & 20.5 
\end{tabular}

\caption{\label{table:bleu}
Automatic evaluation results in 3 test sets. Highest scoring model of each column is marked with bold. BLEU scores were calculated with SACREBLEU toolkit with tokenize ``intl'' ~\citep{DBLP:conf/wmt/Post18}. }
\end{table*}

\subsection{Human evaluation}



It is essential to evaluate the quality of an MT system in order to assess its usability in real-world scenarios. We designed our manual evaluation setups to give us insights on how our MT models would perform in the context of a COVID-domain chatbot. \textit{Uji}, TWB’s chatbot deployed in DRC, is an artificial intelligence-based virtual assistant that allows people to interact in conversations. Uji answers questions, records concerns and feedback in French, Swahili, and Lingala\footnote{\url{https://translatorswithoutborders.org/blog/chatbot-uji-drc/}}. Two main translation tasks involved in the project are: 

\begin{itemize}
    \item Translation of professionally curated responses to common queries related to COVID-19. This content is originally prepared in French and needs to be translated into Swahili. 
    \item Analysis of unclassified queries, complaints and other monitorizable information from Swahili-speaking chatbot users. 
\end{itemize}

For the first use-case, we emulated a translation setup where a linguist would post-edit MT output to generate correct translations. For the second use case, we assessed machine translation output directly. 

\subsubsection{Post-editing evaluation}

In the post-edited translation scenario, translators are given the option to use a machine translation system to translate a document segment by segment. Since MT is prone to error, these translations are then post-edited to ensure quality. One way of evaluating the quality of an MT system is assessing the amount of post-editing effort needed to obtain correct translations~\citep{bentivogli}.

\textbf{Data.} We selected 10 recently prepared response strings originally prepared in French. The responses covered various topics such as vaccinations, negationism, myths, and new virus variants. The strings contain an average of 2.3 sentences and 45 words. 

\textbf{Results.} Table~\ref{table:pe} lists the results that compare automatic and post-edited versions of the translations in terms of BLEU, HTER~\cite{Snover2006ASO} and CHRF~\citep{popovic-2015-chrf}. We found higher than average agreement where our evaluator reports principle errors as missing words, French words appearing, and confusion of plural and singular. An example segment with machine translated (Raw-MT) and post-edited versions (PE-MT) is given below. Corrections made by the translator are marked in red.

\begin{description}
\item [French] {\it Des cas de réinfection du COVID-19 ont été signalés mais sont rares. En général, la réinfection signifie qu'une personne a été infectée (est tombée malade) une fois, s'est rétablie, puis est redevenue infectée plus tard. Sur la base de ce que nous savons de virus similaires, certaines réinfections sont attendues.} 
\item [Raw-MT] {\it Kesi za uambukizi wa COVID-19 zimeripotiwa lakini hazipatikani kwa urahisi. Kwa ujumla, maambukizi inamaanisha kama mtu aliambukizwa (aliugua) mara moja, s' alirudishwa, na kisha akaambukizwa tena baadaye. Kutokana na kile tunachojua kuhusu virusi kama hivyo, maambukizo mengine yangali mbele.}
\item [PE-MT]{\it 
Kesi za \red{maa}mbukizi \red{ya} COVID-19 zimeripotiwa lakini hazipatikani \red{mara nyingi}. Kwa ujumla, maambukizi inamaanisha kama mtu aliambukizwa (aliugua) mara moja, \red{akapona}, na kisha akaambukizwa tena baadaye. Kutokana na kile tunachojua kuhusu virusi kama hivyo, maambukizo mengine yangali mbele.}
\end{description}

\begin{table}
\centering
\begin{tabular}{lll}
\hline
\textbf{BLEU} & \textbf{TER} & \textbf{ChrF} \\
\hline
55.92 & 0.37  & 74.81  \\
\hline
\end{tabular}
\caption{\label{table:pe}
Evaluation results for post-edited translation in the French to Congolese Swahili direction. We used the evaluation setup of \citet{nekoto2020participatory} where TER is computed with \textit{pyter} python package and BLEU and ChrF are computed with \textit{SACREBLEU} with tokenize “none”.}
\end{table}

\subsubsection{Direct Assessment}

We extracted 100 Congolese Swahili strings from user-submitted chatbot conversations. These strings were translated into French using our MT models. We then randomly separated the strings into four unique 25-question surveys, and each survey was shared with two translators. Each translator only responded to one survey. 

Translators were shown the source text along with the target text. These strings were shown independent of each other and not within the context of a wider document or corpora. Translators were then asked three simple questions:

\begin{enumerate}
    \item Does this translation convey the main meaning of the source text? (Yes, Kind of, No)
    \item How good is the previous translation? (0 = very bad; 10 = perfect)
    \item In the case of a low score (0-4) on Question 2, please explain why this is not a good translation. (open ended)
\end{enumerate}

We chose to omit the commonly used \textit{accuracy} and \textit{fluency} metrics~\citep{DBLP:conf/acllaw/GrahamBMZ13} as we feel they are overly academic for our purposes. As our use-case focuses on eliciting a general understanding of user-submitted content, it’s less important for us to have target translations with perfect grammar or strong linguistic fluency. Instead, we’ve focused on comprehension of the key message (Question 1) and general quality (Question 2) to  evaluate the model’s fitness for purpose.

To control for variations in evaluation scales, the results for each string were averaged between the two translators. Generally, there was strong agreement between the translators' rankings, with a median difference of 2 points on the 11-point scale. To avoid discrepancies in individual differences, any string where the translator scores for Question 2 differed by more than 3 points was omitted from the final analysis. In other words, if one translator provided a score of 3 and the other provided a score of 7, results for that string were omitted. We also removed seven strings that contained significant source content in Lingala. The final analysis was based on 71 strings.

Results for Question 1 were also averaged, but using a number-based conversion from the three-tier ranking to break the tie in the 28\% of strings where the two translators responded differently. The results were assigned a value corresponding with their answer (No = 0, Kind of = 2, Yes = 4) and results were averaged between the two translators. So a score of 3 corresponds with a situation where one translator responded “Kind of” and the other responded “Yes.”

\textbf{Data.} All of the strings were real, user-submitted questions related to COVID-19. As these were questions submitted primarily through WhatsApp, the strings were short, averaging 11 words and 63 characters. 

\textbf{Results}. On average, our strings were rated 6.3 out of 10, with a standard deviation of 3.0. It’s important to note that we only defined the bottom and top of the scale, so there was some interpretation required, and different translators could have scored with different personal criteria for how they defined “good.” In general, we did not notice a relationship between quality and string length, but that can partly be explained by the small difference between our shortest (37 characters) and longest strings (144 characters). We did notice a stronger relationship between the quality of the translation and whether the main message was being conveyed. This is not surprising, as we would expect to understand the main message better when the translations are better. Still, the relationship could suggest utility in focusing basic human evaluations on simple questions (like whether the main message was conveyed), as they are easier and quicker for translators to answer.

\begin{figure}
\includegraphics[width=\linewidth]{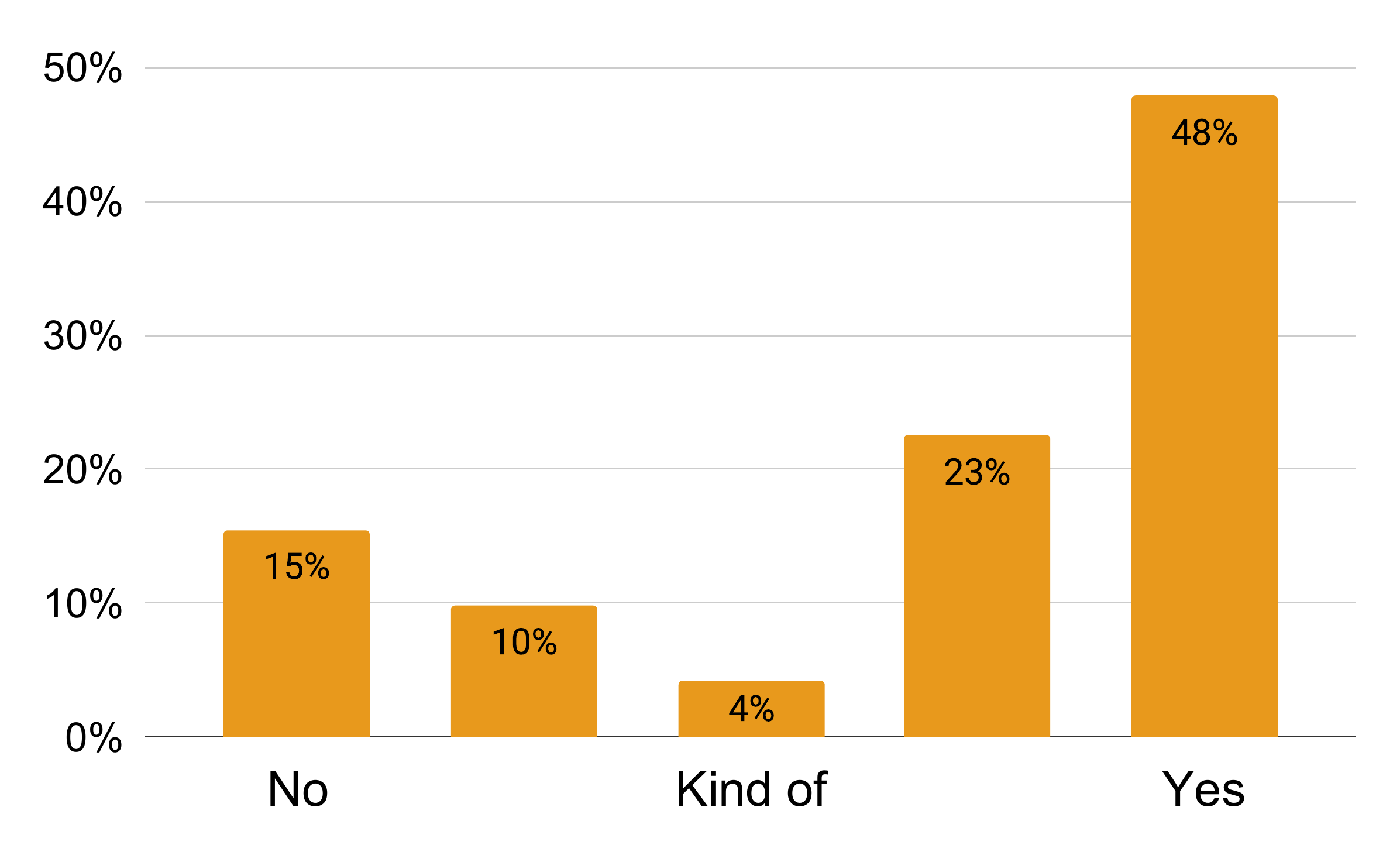}
\caption{Percentage of target translations where the main meaning of the source text was conveyed. Each string was evaluated by two translators and results were averaged.}
\label{fig:meaning_convey}
\end{figure}

In terms of comprehension, our evaluators reported that at least some of the key message of the source text was coming across 75\% of the time (see Figure~\ref{fig:meaning_convey}). In 48\% of the strings, our evaluators both agreed that the key message was being conveyed.

\section{Conclusion and Future Work}

In this work, we have provided a case study of machine translation development in a low-resource setting, Congolese Swahili. We provided detailed experimental results on the use of cross-dialect transfer learning and semi-supervised learning methodologies. Automatic evaluation among various test sets showed an improvement of 1.5 to 2.4 BLEU points in the SWC–FRA direction and 1.4 to 3.5 BLEU points in the FRA–SWC direction through the use of such methods.  

The results of evaluation tests by human translators demonstrate a potential for improving translation workflows in humanitarian relief scenarios. To serve both the general population and translators, we provide a demo translator application serving our models in \url{http://gamayun.translatorswb.org/}. 

With the motivation of promoting NLP research and language technology development for Congolese Swahili, we open-source the resources created in this work. These are: 25,302 clean human-translated general domain sentences and 928,065 synthetic parallel sentences and weights of best-performing models. To ensure the replicability of our work, we also share our development pipeline and test sets under \url{http://github.com/translatorswb/TWB-MT/tree/swc-fra-bidirectional}.

Our current and future work involves integration of our models into our workflows. Our primary objective is giving this tool to TWB’s translator community. We hope to further develop our models with the feedback we receive from linguists and non-professional translators of diverse backgrounds. Also, we aim to achieve impact in humanitarian data collection, such as surveying crisis-affected communities with machine-assisted translation.

\section*{Acknowledgments}

Gamayun kits have been made possible thanks to: Baraka Franck, Joseph Habamungu, Ramazani Katanangwa, Kambale Maboko, Venance Alwende Francois, Kibwe Amani Yao, François Kawalina, Nzay Safina, Jerry Nguwa, Imani Kizungu, Cynthia Ramazani, Ruffin Bindu Ramazani, Liliane Rushukuri, Marie-Claire Akilimali. Thanks to our volunteer evaluators: Diane Ngowire, Joseph Habamungu, Gerson Bwira, Alain Maboko, Akilimali Mirindi, Imani Kizungu. Thanks to Markos Aposporis, Manuel Locria and Paul Warambo for their help during the writing of this paper. This work is funded by Paul G. Allen Family Foundation, Cisco Foundation and Microsoft Corporation.

\bibliography{anthology,eacl2021}
\bibliographystyle{acl_natbib}

\end{document}